\documentclass[11pt]{article} 
\pdfoutput=1
\usepackage{enumerate}
\usepackage{pdfsync}
\usepackage[OT1]{fontenc}
\usepackage{smile}

\usepackage{fullpage}
\usepackage[usenames]{color}
\usepackage[colorlinks,
            linkcolor=red,
            anchorcolor=blue,
            citecolor=blue
            ]{hyperref}
\usepackage{hyperref}
\usepackage{url}
\usepackage{booktabs}
\usepackage{graphicx}
\usepackage{pifont}
\usepackage{multirow}
\usepackage{enumitem}
\usepackage{wrapfig}    
\usepackage{moresize}
\usepackage[dvipsnames]{xcolor}

\definecolor{lightred}{rgb}{1.0, 0.8, 0.8}
\definecolor{lightgreen}{rgb}{0.8, 1.0, 0.8}

\usepackage[textsize=tiny]{todonotes}

\usepackage{color, colortbl}
\definecolor{LightCyan}{rgb}{0.8, 0.9, 1}

\allowdisplaybreaks

\title{\huge Enhancing Multi-Step Reasoning Abilities of Language Models through Direct Q-Function Optimization}

\author
{
   Kaixuan Ji$^*$ \\ {\small  University of California, Los Angeles} \\ {\small \tt kaixuanji@cs.ucla.edu} 
    \\
    \and
   Guanlin Liu\thanks{Equal Contribution}
   \\ {\small ByteDance} \\ {\small \tt guanlin.liu@bytedance.com}
    \and
    Ning Dai \\ {\small  Oregon State University} \\ {\small \tt dain@oregonstate.edu} 
    \\
    \and
    Qingping Yang \\ {\small  ByteDance} \\ {\small \tt qingping95@gmail.com} 
    \\
    \and
    Renjie Zheng \\ {\small ByteDance} \\ {\small \tt renjie.zheng@bytedance.com}
    \and
    Zheng Wu \\ {\small ByteDance} \\ {\small \tt zheng.wu1@bytedance.com} 
    \and
    Chen Dun \\ {\small ByteDance} \\ {\small \tt chen.dun@bytedance.com} 
    \\
    \and
    Quanquan Gu  \\ {\small University of California,  Los Angeles} \\ {\small \tt qgu@cs.ucla.edu}
    \and
    Lin Yan  \\ {\small ByteDance} \\ {\small \tt neil@bytedance.com}
}

\date{}
\begin{document}

\maketitle

\begin{abstract}
Reinforcement Learning (RL) plays a crucial role in aligning large language models (LLMs) with human preferences and improving their ability to perform complex tasks. However, current approaches either require significant computational resources due to the use of multiple models and extensive online sampling for training (e.g., PPO) or are framed as bandit problems (e.g., DPO, DRO), which often struggle with multi-step reasoning tasks, such as math problem solving and complex reasoning that involve long chains of thought. 
To overcome these limitations, we introduce Direct Q-function Optimization (DQO), which formulates the response generation process as a Markov Decision Process (MDP) and utilizes the soft actor-critic (SAC) framework to optimize a Q-function directly parameterized by the language model. The MDP formulation of DQO offers structural advantages over bandit-based methods, enabling more effective process supervision. 
Experimental results on two math problem-solving datasets, GSM8K and MATH, demonstrate that DQO outperforms previous methods, establishing it as a promising offline reinforcement learning approach for aligning language models.
\end{abstract}

\section{Introduction}

Large language models (LLMs) have shown remarkable performances and potentials of a wide range of tasks including dialog generation~\citep{han2024psydial}, general question answering~\citep{alawwad2024enhancing}, code generation~\citep{jimenez2023swe,chen2024teaching}, agents~\citep{wang2024opendevin} and math problem solving~\citep{yumetamath,shao2024deepseekmath}. 
To ensure good performance, one of the key procedures is to align language models with human preferences or task-specific requirements by reinforcement learning (RL)~\citep{bai2022training,touvron2023llama}. Canonically, the alignment training pipeline consists of two stages~\citep{ouyang2022training}. In the first stage, a reward model under the Bradley-Terry model~\citep{bradley1952rank} is trained on human or environment-labeled preference data. Then the language model is trained by online RL algorithms like Proximal Policy Optimization (PPO) \citep{schulman2017proximal} with the reward signals provided by the reward model trained in the first stage.

Despite the good performance achieved, online RL methods usually involve sampling during training, which is both costly and unstable compared to offline methods \citep{Choshen2020On}. These issues are overcome by offline preference learning methods, of which the representative is Direct Preference Optimization (DPO) \citep{rafailov2024direct}. DPO and its follow-ups (e.g., \citet{zhao2023slic,azar2024general}) treat the language model as the policy model and reward model simultaneously and train the model on offline pairwise preference data directly, therefore eliminating the need for a separate reward model. Although simple, direct preference learning has been shown to be effective and efficient in LLM alignment \citep{tunstall2023zephyr}.

However, in practice, sometimes it is hard to acquire pairwise data required by the above methods. This issue becomes particularly severe in the context of math problem solving or code generation when generating high-quality data requires domain-specific expertise \citep{saunders2022self,bowman2022measuring}. This drawback of DPO is circumvented by the recently proposed Direct Reward Optimization (DRO) \citep{richemond2024offline}. DRO formulates the LLM generation task as a bandit and adopts the soft actor-critic (SAC) framework \citep{haarnoja2018soft}, where the advantage is directly parameterized by the language model. Consequently, DRO inherits the advantage of offline policy gradient and gets rid of the dependency on pairwise data. 

\begin{table*}[ht!]
\centering
\caption{A comparison between DQO, offline learning algorithms, including supervised fine-tuning (SFT), reject sampling (RS) \citep{dong2023raft}, DPO \citep{rafailov2024direct}, KTO \citep{ethayarajh2024kto}, DRO~\citep{richemond2024offline} and online algorithm PPO \citep{schulman2017proximal}. DQO enjoys all the benefits listed in the left-most column.}
\resizebox{0.95\columnwidth}{!}
{
\begin{tabular}{l | c | c | c | c | c | c | >{\columncolor{LightCyan}}c}
\toprule
& SFT & RS & DPO & KTO & DRO & PPO & \bf{DQO} \\ 
	\midrule
Free from online sampling during training  & \textcolor{green}{\ding{51}} & \textcolor{green}{\ding{51}} & \textcolor{green}{\ding{51}} & \textcolor{green}{\ding{51}} & \textcolor{green}{\ding{51}} & \textcolor{red}{\ding{55}}  & \textcolor{green}{\ding{51}} \\ 
Learn from negative samples    & \textcolor{red}{\ding{55}}  & \textcolor{red}{\ding{55}}  & \textcolor{green}{\ding{51}} & \textcolor{green}{\ding{51}} & \textcolor{green}{\ding{51}} & \textcolor{green}{\ding{51}} & \textcolor{green}{\ding{51}} \\
Learn from unbalanced samples & \textcolor{red}{\ding{55}}  & \textcolor{red}{\ding{55}}  & \textcolor{red}{\ding{55}}  & \textcolor{green}{\ding{51}} & \textcolor{green}{\ding{51}} & \textcolor{green}{\ding{51}} & \textcolor{green}{\ding{51}} \\ 
Ability to use process supervision  & \textcolor{red}{\ding{55}}  & \textcolor{red}{\ding{55}}  & \textcolor{red}{\ding{55}}  & \textcolor{red}{\ding{55}} & \textcolor{red}{\ding{55}} & \textcolor{green}{\ding{51}} & \textcolor{green}{\ding{51}} \\
\midrule
\end{tabular}
}
\vspace*{-0.2in}
\label{table:motivation}
\end{table*}

Nevertheless, DRO treats the process as a bandit problem, which neglects the intrinsic long-horizon nature of a wide spectrum of tasks that require complex multi-step reasoning like math problem solving and code generation \citep{kang2024mindstar,miaoselfcheck}, where an erroneous reasoning is almost fatal. 
Previous RL research found that if rewards are only provided at the end of the episode, discovering this sparse reward signal is a hard exploration problem and sparse reward functions may not be able to meaningfully distinguish between a wide range of different policies, which makes the training inefficient \citep{Martin18Learning,wilcox2022monte}.
In the meanwhile, recent studies show that signals from process reward models (PRMs) can further boost the performance of language model  \citep{zhang2024rest,lightman2023let}. The positional information of PRM scores usually implies the critical mistakes in the reasoning and therefore provides stronger supervision signals.
However, if the whole generation process is simplified as a single action, the process rewards will be aggregated and the positional information will be lost, implying that DRO cannot efficiently utilize process supervision.



In order to overcome the aforementioned issues, in this paper, we propose Direct $Q$-function optimization (DQO), an offline RL algorithm for LLMs. In DQO, the responding procedure is formulated as a Markov Decision Process (MDP) and our goal is to learn an optimal policy under KL-regularization. Our algorithm adopts the framework of soft $Q$-learning, where the $Q$-function is directly parameterized by the language model. Then both the $Q$-function network and the value network are updated to fit the offline data according to Soft Bellman Equation. The MDP formulation makes DQO a multi-step learning algorithm, and can therefore exploit process reward signals. A holistic comparison of our method and other alignment methods is shown in Table~\ref{table:motivation}. 
Specifically, our contributions are summarized as follows
\begin{itemize}[leftmargin=*]
    \item We propose Direct $Q$-function optimization, or DQO, an offline RL algorithm for LLMs. DQO formulates the LLM reasoning process as an MDP and adopt soft-actor-critic framework to fit the $Q$-value and $V$-value, where the $Q$-function is directly parameterized by the policy network.     Compared to previous methods, DQO learns from offline and negative or unbalanced samples. Moreover, the MDP formulation, compare to bandit, is more favorable for long-horizon tasks and able to exploit process rewards.
    \item We introduce a practical instantiation of DQO, which equips DQO with $\lambda$-return and importance sampling. These techniques stabilize the training process and ensure a good performance.
    \item We empirically compare DQO with a wide range of widely used alignment algorithms on math problem-solving tasks. Experiment results show that DQO outperforms these baselines on both several math-problem-solving datasets. Moreover, as shown by our experiment, when process rewards are available, the performance of DQO can be further boosted, indicating that DQO can benefit from process rewards. 
\end{itemize}

\section{Related Work}

\paragraph{Reinforcement Learning for Language Model Alignment } Aligning language models with human preferences, or reinforcement learning with human feedback (RLHF), dates back to the work of  \citet{wirth2017survey} and \citet{christiano2017deep}. It has been widely applied to a bunch of recent models including GPT-4~\citep{achiam2023gpt}, Gemini~\citep{team2023gemini}, and Llama~\citep{touvron2023llama}, leading to the surprising performance of these models. The alignment procedure usually takes place after supervised finetuning (SFT). In the canonical approaches of RLHF~\citep{ouyang2022training,bai2022training, munosnash}, a reward model is first trained with preference data and then the model is updated with Proximal Policy Optimization (PPO). Another line of works, initiating from Direct Preference Optimization (DPO)~\citep{rafailov2024direct}, include SLiC~\citep{zhao2023slic}, IPO~\citep{azar2024general}, KTO~\citep{ethayarajh2024kto} and so on. These approaches are featured by directly parameterizing the reward models with the language model and then training on offline preference data. Following DPO, one branch of works, including GSHF~\citep{xiong2024iterative}, SPPO~\citep{wu2024self} and INPO~\citep{zhang2024iterative}, adapts DPO or its variant to online samples and iterative training and resulted to state-of-the-art models. 
On the other hand, \citet{richemond2024offline} adapted offline reinforcement learning algorithm to direct preference learning and proposed Direct Reward Optimization (DRO), which combined offline policy learning with a value function learning and updated policy network and value network iteratively. Our work has a similar structure to DRO, but models the language generation as an MDP rather than a bandit, which can take advantage of process supervision during training. 

\paragraph{Multi-step and Long Horizon RL for LLM Alignment } Many tasks for LLMs require LLMs to reason step by step or interact with the environment turn by turn. However, the rewards are usually sparse since they are only provided at the end of a long horizon of reasoning or interactions. In traditional RL literature, one approach towards breaking the curse of lone horizon and sparse reward is to train or estimate an intermediate value function or process reward~\citep{park2024hiql} and use the process reward to guided searching~\citep{torne2023breadcrumbs,zhang2024rest} and RL training. The utilization of process values has also led to better performance for LLM reasoning~\citep{zhang2024rest,lightman2023let}.
Most straightforwardly, \citet{snelloffline} proposed ILQL, which employed implicit Q-learning to train a Q-function network and V-function network. Then, during inference, ILQL uses learned value functions to perturb the log probabilities of the initial policy towards utility-maximizing behavior. 
The success of direct preference learning also stimulates a line of works learning multi-step or multi-turn tasks with direct preference learning. To estimate and utilize process values, ~\citet{chen2024step,lai2024step,xie2024monte} leveraged process reward signals or AI feedback to construct preference pairs for intermediate steps and then updated the model with original DPO. On the other hand, \citet{xiong2024building, shani2024multi} extended the vanilla DPO to accommodate the multi-turn structure. However, these approaches require pairwise data, which might not be available or easy to obtain on some specific occasions. Our work, while following the approach of direct preference learning, eliminates the need for pairwise data and can be boosted by process supervisions. 
After our work, \citet{wang2024offline}, \citet{liu2024improving} also released manualscripts introducing similar algorithms featured with step-wise actor-critic framework and direct preference learning.


\section{Preliminaries}

In this section, we introduce the foundational concepts and notations that underpin our proposed algorithm. We first review the basic framework of modeling language generation as a reinforcement learning task, followed by a KL-regularized reinforcement learning objective.

\paragraph{Modeling Language Generation as Token-Level MDP } 
Reinforcement Learning (RL) is concerned with learning a policy that maximizes the cumulative reward for an agent interacting with an environment. 
In this work, we formalize language generation tasks as a Markov decision process (MDP). 
We denote prompt as $x$ and a response to the prompt as $y$, which can each individually be broken down into a sequence of tokens, for example, $x = (x_0, \dots, x_m)$, from a fixed discrete vocabulary $\mathcal{A}$.
We define the token-level MDP as a tuple $\mathcal{M} = (\mathcal{S}, \mathcal{A}, \mathbb{P}, H, r, d_0, \omega)$. In the defined MDP, $\mathcal{S}$ is the space of the state consisting of all tokens generated so far, i.e., $s_t = (x_0, \dots, x_m, y_1, \dots, y_{t-1})$. The action space $\mathcal{A}$ is the fixed discrete vocabulary. The dynamics $\mathbb{P}$ are the deterministic transition model between tokens, i.e., $\mathbb{P}(s_{t+1}|s_t,a) = 1$ for $s_t = (x_0, \dots, x_m, y_1, \dots, y_{t-1})$, $a = y_{t}$ and $s_{t+1} = (x_0, \dots, x_m, y_0, \dots, y_t)$\footnote{For notational simplicity, we ignore the case that LLM can call an external tool. If the tool does not introduce randomness, the state transitions are also deterministic. Even if the state transitions are random, we can use samples to approximate the state transition probabilities.}. The generation process will terminate once the terminal action $\omega$ (usually end-of-sentence token) is taken or reaches the maximum horizon length $H$. 
The reward function $r(s,a)$ provides scalar feedback for the agent’s performance after taking action $a$ in state $s$. In RLHF, the reward function is usually learned from human feedback over preferences or given by a series of rules depending on the specific tasks. The initial state distribution $d_0$ is a distribution over prompts $x$, where an initial state $s_0$ is comprised of the tokens from $x$.

\paragraph{KL-Regularized Reinforcement Learning Objective } We follow previous works~\citep{rafailov2024direct, richemond2024offline} and consider the KL-regularized RL problem, of which the objective function $J(\pi)$ is defined as follows:
\begin{align}
    J(\pi) = \mathbb{E}_{\pi, d_0} \bigg[\sum_{h=1}^H  \bigg( r(s_h, a_h) - \beta \log \frac{\pi(a_h | s_h)}{\pi_{\text{ref}}(a_h | s_h)}  \bigg) \bigg], \label{eq:objective}
\end{align}
where $H$ is the total number of decision steps, $s_0$ is a prompt sampled from the dataset, $r(s_h, a_h)$ is the token-level reward from the reward function, $\beta$ is the coefficient controlling the magnitude of KL-regularization and $\pi_{\text{ref}}$ is the initialisation policy. Our goal is to approximate the optimal KL-regularized policy $\pi^*$ given by:
\begin{align*}
   \pi^* =  \argmax_\pi J(\pi).
\end{align*}
 
In classic RLHF and most LLM-related tasks, the reward is sparse and is only applied at the terminal action $\omega$, i.e. the end-of-sentence token \texttt{<eos>}. However, our structure is flexible enough to incorporate both dense and sparse rewards from ruled-based reward models, turn-level reward models, process-supervised reward models (PRM), or just outcome-supervised reward models.

We consider rewriting our objective function~\eqref{eq:objective} under the framework of max-entropy reinforcement learning. Specifically, we decompose the KL-regularization term $\mathrm{KL}(\pi(\cdot | s_h) \| \pi_{\text{ref}}(\cdot | s_h))$ into cross-entropy and entropy, leading to the following max-entropy-RL objective:
\begin{align*}
J(\pi) =  \mathbb{E}_{\pi, s_0 \sim d_0} \bigg[ \sum_{h=1}^H  \Big(\overline{r}(s_h, a_h)+ \beta \mathcal{H}\big(\pi(\cdot | s_h)\big) \Big) \bigg], 
\end{align*}
where $\mathcal{H}(\pi(\cdot | s_h)) = -\mathbb{E}_{a \sim \pi} \log \pi(a|s_h)$ denotes the entropy of the policy at state \( s_h \) and the KL-regularized reward $\overline{r}$ is defined as 
$\overline{r}(s, a) = \beta \log \pi_{\text{ref}}(a | s) + r(s, a)$.
Such max-entropy RL problem enjoys a well-known closed-form solution \citep{haarnoja2018soft} as follows:
\begin{align} 
    \pi^* (a | s) = \exp{\bigg( \frac{Q^*(s, a) - V^*(s)}{\beta} \bigg)}, \text{ or, } Q^*(s, a) = \beta \log \pi^*(a | s) + V^*(s) \label{eq:optpi},
\end{align}
where $Q^*$ and $V^*$ are shorthands for the soft $Q$-function $Q^{\pi^*}$ and soft $V$-function $V^{\pi^*}$ induced by optimal policy $\pi^*$. Here, the soft $V$-function of a policy $\pi$ is defined as
\begin{align*}
V^{\pi}(s_h) &= \mathbb{E}_{\pi} \Bigg[ \sum_{t=h}^{H} \Big(\overline{r}(s_t, a_t) + \beta \mathcal{H}\big(\pi(\cdot | s_t)\big) \Big) \Bigg],
\end{align*}
and the soft $Q$-function of a policy $\pi$ is defined as
\begin{align*}
    Q^{\pi}(s_h, a) &= \overline{r}(s_h, a) + \mathbb{E}_{s_{h+1}} \big[ V^{\pi} (s_{h+1}) \big].
\end{align*}

Equation~\eqref{eq:optpi} reveals that the optimal policy \( \pi^* \), soft $Q$-function \( Q^* \), and soft V-function \( V^* \) are interdependent, which means that knowing any two of them allows us to compute the third one.

\section{Direct $Q$-function Optimization (DQO)}

\subsection{The DQO objective}\label{sec:dqo-obj}

We adopt the Soft Actor-Critic (SAC) learning framework to learn the state value function $V$ and state-action value function $Q$. In SAC, the $Q$-function and $V$-function, which are parameterized by $\theta$ and $\phi$ respectively, are updated by minimizing the following squared Bellman residuals: 
\begin{align}
    & L_V(\phi) = \mathbb{E}_{(s_h, a_h, s_{h+1}) \sim \cD} \big[ \big( V_\phi(s_h) -  Q_\theta(s_h, a_h) + \beta \log \pi_{\theta}(a_h|s_h) \big)^2 \big] \label{eq:SACV}, \\
    & L_Q(\theta)  = \mathbb{E}_{(s_h, a_h, s_{h+1}) \sim \cD} \big[ \big( Q_\theta(s_h, a_h) -  \overline{r}(s_h, a_h) - V_\phi(s_{h+1}) \big)^2 \big], \label{eq:SACQ}
\end{align}
where $\mathcal{D}$ is the distribution of previously sampled states and actions and $\theta$ is the parameter of $Q$-function (which is essentially the policy, or the LLM, in DQO).
For simplicity of notations, we always set $V_{\phi}(s_{H+1})=0$ for all $\phi$ and $s_{H+1}$. 
Learning the parameters by optimizing over~\eqref{eq:SACV} and~\eqref{eq:SACQ} requires three sets of parameters, the parameters for $Q$-function, $V$-function and the policy. Inspired by recent advancement in direct preference learning, we eliminate the requirement of $Q$-function parameter as follows. First, to eliminate the $Q_{\theta}(s_h, a_h)$ in~\eqref{eq:SACV}, we consider the soft Bellman equation:
\begin{align*}
    Q^{\pi}(s_h, a_h) = \overline{r}_h(s_h, a_h) + V^{\pi}(s_{h+1}),
\end{align*}
where under deterministic transition $s_{h+1}=\text{concat}(s_h, a_h)$. Consequently, we can rewrite~\eqref{eq:SACV} and obtain the loss for the value function $V_{\phi}$ in the following form without $Q_\theta$:
\begin{align}
L_V(\phi) = \mathbb{E}_{(s_h, a_h, s_{h+1}) \sim \cD} \big[ \big( V_\phi(s_h) - \overline{r}(s_h, a_h) - V_{\phi}(s_{h+1}) + \beta \log \pi_{\theta}(a_h|s_h) \big)^2 \big], \label{eq:SACV-rewrite}
\end{align}

Now we come to the loss for $Q$-function~\eqref{eq:SACQ}. 
As shown in~\eqref{eq:optpi}, the optimal policy $\pi^* $, optimal $Q$-function $Q^*$, and optimal value function $V^*$ are connected. Inspired by this, we parameterize the $Q$-value-network with the policy as follows
\begin{align}\label{eq:reparam-q}
    Q_\theta(s_h, a_h) = \beta \log \pi_\theta(a_h | s_h) + V_\phi(s_h),
\end{align}
where $\pi_{\theta}(\cdot | \cdot)$ is the policy network, or the language model in the context of LLM reasoning. 
By plugging equation~\eqref{eq:reparam-q} into equation~\eqref{eq:SACQ}, we can rewrite the Q-function target as
\begin{align}
    L_\pi(\theta)  =   \mathbb{E}_{(s_h, a_h, s_{h+1}) \sim \cD} \big[ \big( V_\phi(s_h) + \beta \log \pi_\theta(a_h | s_h) - \overline{r}(s_h, a_h) - V_\phi(s_{h+1}) \big)^2 \big] \label{eq:SACQ-rewrite},
\end{align}
\color{black}


Compared to the original objective~\eqref{eq:SACV} and~\eqref{eq:SACQ}, which use an independent model to parameterize $Q$-value and learn the policy from the optimal $Q$-function $Q^*$, with our objective~\eqref{eq:SACV-rewrite} and~\eqref{eq:SACQ-rewrite}, we directly infer the policy from the $Q$-function by parameterizing it with $\pi$. Therefore, we dub our algorithm as Direct $Q$-function optimization (DQO).

\begin{remark}\label{remark:necessity-is}
The loss function~\eqref{eq:SACV} is minimized when $V_\phi(s_h) = \mathbb{E}_{(s_h, a_h, s_{h+1}) \sim \cD}[Q_\theta(s_h, a_h) - \beta\log \pi_{\theta}(a_h|s_h)]$, which agrees with the definition $V_\phi(s_h) = \EE_{a\sim \pi}[Q_\theta(s_h, a_h) - \beta\log \pi_{\theta}(a_h|s_h)]$ only when $\cD$ is generated from the current policy $\pi$. However, this does not hold when $\mathcal{D}$ is composed of pre-generated offline data. Therefore, importance sampling is required to properly re-weight the offline data to mitigate this misalignment. We defer the detailed discussion to Section~\ref{sec:IS}.
\end{remark}

\begin{remark}
In our formulation of DQO, we consider generating a single token as an action. If we consider generating the whole utterance as a single action and setting the horizon length $H=1$, then equation~\eqref{eq:SACV-rewrite} and~\eqref{eq:SACQ-rewrite} degenerate to the loss used by {DRO}~\citep{richemond2024offline}. This means that DRO can be viewed as a special case of the learning framework of DQO.
\end{remark}


\subsection{Mitigating Bias with $\lambda$-Return}

One-step temporal difference (TD) errors have high bias and perform poorly when the value function is not well-initialized, resulting in inefficient learning. To address this, we incorporate $\lambda$-return~\citep{schulman2015high} to improve the updates for $Q$-function and $V$-function. By definition, we know that $V^{\pi}(s_h)$ is the sum of reward gained by next $n$ actions and $V^{\pi}(s_{h+n})$, or formally, 
\begin{align*}
    V^\pi(s_h) = \mathbb{E}_{\tau \sim \pi} \big[ V^\pi(s_{h+n}) + \Delta_{h, n}^{\pi}(\tau)\big],
\end{align*}
where $\tau$ is a trajectory and 
\begin{align*}
    & \Delta_{h, n}^{\pi}(\tau)  =\sum_{l=0}^{n-1} \bigg( r(s_{h+l}, a_{h+l}) - \beta \log \frac{ \pi(a_{h+l} | s_{h+l})}{\pi_{\text{ref}}(a_{h+l} | s_{h+l})} \bigg).
\end{align*}
Given a trajectory $\tau = \{s_0,a_0,\cdots,s_H,a_H\}$, we use the empirical samples to estimate the $n$-step return and define the empirical $n$-step return as:
\begin{align*}
    G_{\phi,\theta}^{(n)}(s_h) = V_\phi(s_{h+n}) + \Delta_{h,n}^{\pi}(\tau) .
\end{align*}
It is worth noticing that $G^{(1)}_{\phi, \theta}(s_h)$ is exactly the target in~\eqref{eq:SACV-rewrite}. Now we are able to define $\lambda$-return, which is the weighted average of all $n$-step returns:
\begin{align*}
G_{\phi,\theta}^\lambda(s_h) = \begin{cases}
    (1 - \lambda) \sum_{n=1}^{H-h} \lambda^{n-1} G_{\phi,\theta}^{(n)}(s_h), & \text{if } \lambda < 1 \\
    G_{\phi,\theta}^{(H-h)}(s_h), & \text{if } \lambda=1
\end{cases} .
\end{align*}
We replace the target for value updates in~\eqref{eq:SACV-rewrite} from one-step return $G^{(1)}_{\phi, \theta}(s_h)$ to $\lambda$-return $ G_{\overline{\phi},\overline{\theta}}^\lambda(s_{h})$, where $\overline{\phi}$ and $\overline{\theta}$ is the copy of $\phi$ and $\theta$ but are not counted into the back-propagation gradients. Now we have the loss function for the value network as follows:
\begin{align}
    L_V(\phi) = \mathbb{E}_{s_h \in \mathcal{D}} \Big[ \big(G_{\overline{\phi},\overline{\theta}}^\lambda(s_h) - V_\phi(s_h)\big)^2 \Big]. \label{eq:sacv-lambda}
\end{align}
Similarly, we also use the $\lambda$-return $G_{\overline{\phi},\overline{\theta}}^\lambda(s_{h+1})$ to substitute the target $V_{\phi}(s_{h+1})$ in~\eqref{eq:SACQ-rewrite}. The new loss for $Q$-function (which is parameterized by $\pi_{\theta}$) with $\lambda$-return is:
\begin{align}
L_{\pi}(\theta)  &= \mathbb{E}_{(s_h, a_h, s_{h+1}) \sim \cD} \big[ \big( V_\phi(s_h) + \beta \log \pi_\theta(a_h | s_h) -\overline{r}(s_h, a_h) - G_{\overline{\phi},\overline{\theta}}^\lambda(s_{h+1}) \big)^2 \big] \notag \\
& = \mathbb{E}_{(s_h, a_h, s_{h+1}) \sim \cD} \bigg[ \bigg( V_\phi(s_h) + \beta \log \frac{\pi_\theta(a_h | s_h)}{\pi_{\text{ref}}(a_h | s_h)}  - r(s_h, a_h) - G_{\overline{\phi},\overline{\theta}}^\lambda(s_{h+1}) \bigg)^2 \bigg] \label{eq:sacq-lambda}.
\end{align}

\subsection{Reweighting Offline Data with Importance Sampling} \label{sec:IS}
As we have mentioned in Remark~\ref{remark:necessity-is}, the optimization objective~\eqref{eq:SACV} and~\eqref{eq:SACQ} agree with Bellman equation only when the dataset is online sampled from $\pi_{\theta}$. However, in the offline setting, the data is pre-collected from $\pi_{\mathrm{ref}}$, causing a distributional shift between the behavior policy $\pi_{\mathrm{ref}}$ which generated the data, and the target policy $\pi_{\theta}$. 
In order to mitigate this mismatch, 
we employ importance sampling to reweight the offline data to match the distribution of trajectories generated by the current policy. It enables us to leverage offline datasets in an online RL framework.

Let $\mu$ represent the behavior policy under which the offline data $\mathcal{D}$ was generated and $\pi$ be the current online policy. For any function of trajectory $f$, we have 
\begin{align*}
    \mathbb{E}_{\tau \sim \pi}[f(\tau)] &= \mathbb{E}_{\tau \sim \mathcal{D}}\bigg[ \frac{\pi(\tau | s_1)}{\mu(\tau | s_1)} f(\tau)\bigg] ,
\end{align*}
where the probability of a trajectory $\tau$ under $\mu$ and $\pi$ are computed as follows:
\begin{align*}
\mu(\tau | s_1) = \prod_{h=1}^H \mu(a_h | s_h), \quad \pi(\tau | s_1) = \prod_{h=1}^H \pi(a_h | s_h),
\end{align*}
when the transition $\mathbb{P}$ is deterministic. 
We truncate the importance sampling rate to avoid gradient explosion caused by extreme values and obtain the final ratio $w(\tau) = \min\{ \pi(\tau|s_1)/\mu(\tau|s_1), e\}$.
Now by setting $\mu=\pi_{\mathrm{ref}}$ and $\pi=\pi_{\theta}$, and applying the importance ratio $w(\tau)$ to the loss functions~\eqref{eq:sacv-lambda} and~\eqref{eq:sacq-lambda}, we obtain our final loss functions for offline learning.
\begin{align*}
& L_V(\phi) = \mathbb{E}_{\tau \sim \mathcal{D}} \Big[ w(\tau) \cdot \big(G_{\overline{\phi},\overline{\theta}}^\lambda(s_h) - V_\phi(s_h)\big)^2 \Big], \\
& L_{\pi}(\theta) = \mathbb{E}_{\tau \sim \cD} \bigg[ w(\tau)\bigg( V_\phi(s_h) + \beta \log \frac{\pi_\theta(a_h | s_h)}{\pi_{\text{ref}}(a_h | s_h)}  - r(s_h, a_h) - G_{\overline{\phi},\overline{\theta}}^\lambda(s_{h+1}) \bigg)^2 \bigg] \label{eq:sacq-lambda}.
\end{align*}

When computing the gradient of the loss, the importance sampling weight $w(\tau)$ is not involved in the gradient computation. 


\section{Experiments}

In this section, we conduct extensive experiments to demonstrate the effectiveness of our proposed method. Moreover, we show that our method can be further augmented by utilizing process rewards.

\begin{table*}[ht!]
\centering
\caption{Experiment results on Qwen2.5-QwQ model. We use \textbf{bold} for the best and \underline{underline} for the second best. DQO significantly improves the base model's performance and the improvement surpass all the baselines. } 
\vspace*{0.1in}
\begin{tabular}{ l | c | c | c | c | c}
\toprule
Dataset & \multirow{2}{*}{AIME24}  & \multicolumn{2}{c|}{GSM8K}  & \multicolumn{2}{c}{MATH}  \\
Decoding    &   & Greedy   & Sample    & Greedy     & Sample  \\
\midrule
Qwen2.5-QwQ &  14.33  & 76.57 & 69.32$_{\pm 0.51}$ & 58.56 & 48.84$_{\pm 0.41}$   \\
\midrule
Qwen2.5-QwQ + RS   & \underline{23.11}     & \underline{82.03} &\underline{81.09$_{\pm 0.51}$} & \underline{71.40}   & \underline{66.07$_{\pm 0.56}$}    \\
Qwen2.5-QwQ + DPO  & 17.44  & \underline{82.03} &  71.95$_{\pm 0.15}$   & 68.16  & 53.12$_{\pm0.37}$          \\
Qwen2.5-QwQ + KTO  &  15.55   & 78.62   & 69.49$_{\pm0.11}$  & 61.26  & 49.68$_{\pm 0.68}$   \\
Qwen2.5-QwQ + DRO & 17.00   & 79.91 & 72.58$_{\pm 0.88}$ & 66.92 & 55.42$_{\pm 0.71}$  \\
\midrule
\rowcolor{LightCyan} Qwen2.5-QwQ + DQO &\textbf{25.33}   & \textbf{83.01} & \textbf{81.95$_{\pm 0.66}$}  & \textbf{74.22}  & \textbf{67.34$_{\pm0.25}$} \\ 
\bottomrule
\end{tabular}
\label{tab:result-qwen25}
\end{table*}

\begin{table*}[ht!]
\centering
\caption{Experiment results for gemma-1.1-it-7B model. We use \textbf{bold} for the best performance and \underline{underline} for the second best performance. DQO significantly improves the 
base model's performance. On GSM8K, DQO surpasses all other baselines by a significant margin. On MATH dataset, DQO achieves a comparable performance with DRO when doing greedy decoding and outperforms all the baseline when doing sampling at inference.} 
\vspace*{0.1in}
\begin{tabular}{l | c | c | c | c }
\toprule
Dataset                 & \multicolumn{2}{c|}{GSM8K} & \multicolumn{2}{c}{MATH}  \\
Decoding      & Greedy  & Sample  & Greedy   & Sample \\
\midrule
gemma-1.1-it-7B    & 39.65    & 37.89$_{\pm 1.02}$ & 17.04   & 16.14$_{\pm 0.21}$ \\
\midrule
gemma-1.1-it-7B + SFT  & 53.45    & 46.14$_{\pm 1.07}$  & 21.64 & 18.84$_{\pm 0.47}$ \\
gemma-1.1-it-7B + RS        & 53.60    & 53.17$_{\pm 0.94}$     & 21.74   & 20.77$_{\pm 0.26}$  \\
gemma-1.1-it-7B + DPO   & \underline{63.46}    & 62.76$_{\pm 0.48}$ & 23.18   & 23.44$_{\pm 0.30}$  \\
gemma-1.1-it-7B + KTO   & 50.49    & 49.29$_{\pm 0.74}$  & 18.56   & 18.58$_{\pm 0.17}$ \\
gemma-1.1-it-7B + DRO  & 62.92    & \underline{63.00$_{\pm 0.92}$}  & \underline{24.56}      & \underline{24.10$_{\pm 0.37}$} \\
\midrule
\rowcolor{LightCyan} gemma-1.1-it-7B + DQO   & \textbf{64.51} & \textbf{64.00$_{\pm 0.37}$ } & \textbf{24.90}  & \textbf{24.84$_{\pm 0.29}$}   \\
\bottomrule
\end{tabular}
\label{tab:result-gemma}
\end{table*}

\subsection{Models and Dataset Constructions}
We select \textit{Qwen2.5-Math-7B}\footnote{https://huggingface.co/Qwen/Qwen2.5-Math-7B}~\citep{yang2024qwen25} and \textit{gemma-1.1-it-7B}\footnote{https://huggingface.co/google/gemma-1.1-7b-it} (Gemma) \citep{team2024gemma} as our base model. To enhance the long chain-of-thought ability, we further fine-tune \textit{Qwen2.5-Math-7B} with  4096 randomly sampled trajectories from a public SFT dataset QwQ-LongCoT-Verified-130K\footnote{https://huggingface.co/datasets/qingy2024/QwQ-LongCoT-Verified-130K}, in which the responses were generated using QwQ-32B-Preview. Then we obtain the fine-tuned model denoted as Qwen2.5-QwQ.
Leveraging Qwen2.5-QwQ's higher proficiency, we train it on more challenging Olympiad-level problems from NuminaMATH~\citep{numina_math_datasets} (Numnina), while Gemma is trained on GSM8K~\citep{cobbe2021gsm8k} and MATH~\citep{hendrycks2021measuring} to better align with its skill level and scope.
For each base model and problem set, we rollout the base model and sample 20 responses for each problem in the training set and then label all these responses as positive and negative responses. These labeled responses are then utilized to construct our training set for each baselines and we refer the readers to Appendix~\ref{app:dataset} for detailed procedure of the construction.

\subsection{Baselines and Evaluation}

We select SFT, reject sampling (RS), DPO, KTO and DRO as our baselines and implement our method based on HybridFlow~\citep{sheng2024hybridflow}. Please refer to Appendix~\ref{app:baseline-hyper} for th  training details of the baselines and our method. 

We employ three widely used dataset to evaluate the models, namely GSM8K~\citep{cobbe2021gsm8k}, MATH \citep{hendrycks2021measuring} and AIME24\footnote{https://huggingface.co/datasets/di-zhang-fdu/AIME\_1983\_2024}, where
AIME24 refers to American Invitational Mathematics Examination 2024.
Since problems from AIME24 is much harder than those from GSM8K and MATH, we only applt AIME24 to Qwen2.5-QwQ-based models. We consider two different decoding strategies for GSM8K and MATH, greedy decoding and sampling. We set the sampling parameters to the same as when we generated the training corpus. 
For each prompt in the dataset, we sample with 5 different seeds and report the mean and standard deviation of the performance. For AIME24, we follow the convention and sample 30 responses for each problem and report the average passing rate.

\subsection{Empirical Results and Case Studies}

Our main results are shown in Table~\ref{tab:result-qwen25} and Table~\ref{tab:result-gemma} respectively for Qwen2.5-QwQ and Gemma. From Table~\ref{tab:result-qwen25}, we see that all the methods improve the performance of the base models by a significant margin. Particularly, on GSM8K, DQO improves the performance from 76.57\% to 83.01\% for greedy decoding and 69.32\% to 81.95\% for sampling. This improvement surpasses RS and other baselines by a margin of at least 0.98\% for greedy decoding and 0.88\% for greedy generation.
On MATH and AIME24, we observe a much more significant performance improvement over the baselines. On MATH dataset, as for greedy decoding, DQO improves the performance from 58.56\% to 74.22\% and such improvement surpasses all other baselines by a margin of more than 2.82\%. As for sampling, DQO also surpasses all other baselines by a margin of more than 1.27\% on MATH and 2.22\% on AIME24. These results indicate the superior performance of DQO over all other baselines on Qwen2.5-QwQ model.

In terms of Gemma results, we see that DQO also enjoys considerable advantages. As demonstrated in Table~\ref{tab:result-gemma}, we see that all considered methods result in significant improvement. Specifically, on GSM8K, DQO improves the base model's performance by a margin of 24.86\% for greedy decoding and 26.11\% for sampling. These results surpass the improvement obtained by DPO by margins of 1.05\% as for greedy decoding and is comparable to the performance of DRO when sampling. The advantage is even more compared to other baseline methods. On the MATH dataset, we see that DQO also improves the model's performance by a prominent margin of 7.86\% and 8.70\% for greedy generation and sampling, respectively. This improvement slightly surpasses DRO and surpasses other baseline methods by a margin of at least 2.28\% for greedy decoding and 2.56\% for sampling. In summary, DQO results in promising improvement over the base models under all the scenarios and outperforms all our baseline methods.

\begin{table*}[ht!]
\centering
\Huge
\caption{A case study of learned $V$-function. We select a prompt-response pair from the GSM8K training set and split the response into nine segments by double line breaks. We extract the $V$-function value at each line break. The model makes a mistake at the beginning of the fifth segment (where \texttt{\$\$x=y\$\$} is erroneous). The model predicts high $V$-values from the first segment to the fourth segment. After that, the predictions drop significantly. The value trends preciously reflect the correctness of the reasoning and thus can serve as an informative process supervision.}
\vspace*{0.1in}
\resizebox{0.95\columnwidth}{!}{
\begin{tabular}{l | l | c | c}
\toprule
&  \textbf{Text} & \textbf{Process Value} & \textbf{Correctness} \\ \midrule
\textbf{Prompt} &  
\parbox[t]{25cm}{
Jorge has an equal number of baseball cards as Matias, who has 6 fewer cards than Carlos.
If Carlos has 20 baseball cards, what is the total number of baseball cards the three have? \\
Please reason step by step, and produce a final answer following 4 '\#', like '\#\#\#\# 0'. \vspace{15pt}
}
 & - & -\\ \midrule
\textbf{Response} & \colorbox{lightgreen}{\parbox[b]{25cm}{\texttt{**Step 1: Identify the information provided**} \\}} & 0.6016 & \textcolor{Green}{\ding{51}} \\ 
  & \colorbox{lightgreen}{\parbox[b]{25cm}{ \texttt{- Jorge has an equal number of baseball cards as Matias.\\
     - Matias has 6 fewer cards than Carlos.\\
     - Carlos has 20 baseball cards.} \\}}
     & 0.6719 &  \textcolor{Green}{\ding{51}} \\
  & \colorbox{lightgreen}{\parbox[b]{25cm}{ \texttt{**Step 2: Translate the information into equations**} \\}}
     & 0.7969 & \textcolor{Green}{\ding{51}}  \\
  & \colorbox{lightgreen}{\parbox[b]{25cm}{ \texttt{Let x be the number of baseball cards Jorge and Matias each have. \\
Let y be the number of baseball cards Carlos has.} \\}}
     & 0.6680 & \textcolor{Green}{\ding{51}}  \\
  &  \colorbox{lightred}{ \parbox[b]{25cm}{ \texttt{\$\$x = y\$\$ \$\$y = x + 6\$\$ \$\$y = 20\$\$}\\} }
     & 0.1016 & \textcolor{red}{\ding{55}}  \\
  &  \colorbox{lightred}{\parbox[b]{25cm}{ \texttt{**Step 3: Substitute the equations to find x**}\\} }
     & 0.0013 & \textcolor{red}{\ding{55}}  \\
  & \colorbox{lightred}{ \parbox[b]{25cm}{ \texttt{\$\$x = x + 6\$\$ \$\$0 = 6\$\$ This is a contradiction, so the given information is not consistent.}\\}} 
     & 0.4688 & \textcolor{red}{\ding{55}}  \\
  & \colorbox{lightred}{ \parbox[b]{25cm}{ \texttt{**Therefore, there is no solution to the problem as the given constraints are contradictory.**}\\} }
     & {0.1768} & \textcolor{red}{\ding{55}}  \\ 
    & \parbox[b]{25cm}{ \colorbox{lightred}{ \texttt{\#\#\#\# 0 }} }    & - & \textcolor{red}{\ding{55}}  \\ 
\bottomrule
\end{tabular}
}
\label{table:example}
\end{table*}

\paragraph{Case Study } In DQO, we formulate the LLM reasoning process as a Markov Decision Process rather than bandit and incorporate a trainable value function for each state to provide progress supervision. Here we provide a prompt-response pair from Gemma-generated GSM8K training set to demonstrate how learned $V$-value provide progress supervision. We extract the $V$-function value at each double-break-line and manually check the correctness of each line of the response. The results are presented in Table~\ref{table:example}. The response makes a mistake at the beginning of the fifth segment (where \texttt{\$\$x=y\$\$} is erroneous). Correspondingly, the trained value model predicts high $V$-values from the first segment to the fourth segment. After that, the predictions drop significantly, which indicates that the value model introduced in DQO can serve as an informative process supervision. 
We refer the readers to Appendix~\ref{app:case-study} for additional cases.

\subsection{Ablation Studies}

In this section, we perform ablation studies to illustrate the role of two key aspects, $\lambda$-return and importance sampling.

\begin{table}[ht!]
\centering
\caption{The impact of $\lambda$-return on Gemma. When decreasing $\lambda$ from 1.0 to 0.95, we observe a significant performance dropping more than 4.31\% on GSM8K and 2.30\% on MATH. } 
\vspace*{0.1in}
\begin{tabular}{ l | c | c | c | c}
\toprule
   Dataset  & \multicolumn{2}{c|}{GSM8K} & \multicolumn{2}{c}{MATH} \\
  Decoding       & Greedy  & Sample  & Greedy   & Sample \\
\midrule
 $\lambda=0.95$   & 60.20    & 59.21     & 22.60    & 22.26    \\
$\lambda=1.0$     & \textbf{64.51} & \textbf{64.00}  & \textbf{24.90} & \textbf{24.84}    \\
\bottomrule
\end{tabular}
\vspace*{0.1in}
\label{tab:ablation-gae-return}
\end{table}

\paragraph{$\lambda$-return }
In order to demonstrate the impact of $\lambda$-return, we vary the value of $\lambda$ and evaluate the training results on Gemma. Empirically, we find that the best performance is obtained at $\lambda=1$ and quickly degenerates when decreasing $\lambda$. Therefore we pick $\lambda=0.95$ to make the comparison. The results are shown in Table~\ref{tab:ablation-gae-return}. When switching $\lambda$ to 0.95, we observe that the performance on GSM8K decreases by a margin of more than 4.31\% for greedy generation and almost 5\% for sampling. The results on MATH demonstrate a similar pattern and the performances of $\lambda=0.95$ dropped by a margin of 2.30\% on both inference strategies. The results indicate that $\lambda$-return plays a crucial role in stabilizing the training process.

\begin{table}[ht]
\centering
\caption{The impact of importance sampling on both $Q$-function loss and $V$-function loss, where we use $Q_{\text{w/ IS}}$ as shorthand for $Q$-function loss with importance sampling, $Q_{\text{w/o IS}}$ for $Q$-function loss without importance sampling and similarly for $V$-function loss. When training without an importance sampling ratio on $Q$-function loss, the performances degenerate significantly. When keeping the importance ratio only on $Q$-function loss, there is also a moderate performance loss on MATH.} 
\vspace*{0.1in}
\renewcommand{\arraystretch}{0.85}
\begin{tabular}{l | c | c | c | c | c}
\toprule
\multicolumn{2}{l |}{Dataset}   & \multicolumn{2}{c|}{GSM8K} & \multicolumn{2}{c}{MATH} \\
\multicolumn{2}{l |}{Decoding}       & Greedy  & Sample  & Greedy   & Sample \\
\midrule
\multirow{2}{*}{$Q_{\text{w/o IS}}$} & $V_{\text{w/o IS}}$& 58.68      &  60.20   &21.96     &  22.68   \\
& $V_{\text{w/ IS}}$ & 56.03      &  56.48   & 20.82     &  20.94  \\
\midrule
\multirow{2}{*}{$Q_{\text{w/ IS}}$} & $V_{\text{w/o IS}}$ &  63.53    &  \textbf{64.06}    &  22.28  &   23.18   \\
& $V_{\text{w/ IS}}$ & \textbf{64.51} & 64.00  & \textbf{24.90} & \textbf{24.84}    \\
\bottomrule
\end{tabular}
\label{tab:ablation-importance}
\end{table}

\paragraph{Importance Sampling } To demonstrate the impact of the importance sampling ratio in DQO, we train DQO on Gemma without the importance sampling ratio for $Q$-function loss, $V$-function loss, or both. We present the results in Table~\ref{tab:ablation-importance}. The results show that, without adding importance sampling, the performance will be significantly deteriorated. Specifically, on the GSM8K dataset, when importance sampling is not introduced to $Q$-function loss, the performance degenerates by a margin over 3.80\%. Similarly, on the MATH dataset, we see that when we exclude the importance sampling ratio from $Q$-function loss, the performance decreases by a margin over 2.16\%. When we keep the importance sampling ratio only on $Q$-function loss, the performance on GSM8K almost maintains, but we still see a moderate performance loss on MATH. These results show that the importance sampling ratio, on both $Q$-function and $V$-function loss, plays an important role in DQO training.


\subsection{DQO with Process Score}

In this section, we show that when process scores are available, the performance of DQO can be further improved. We synthesize a process score for each response in the training set. For each response, we split it to segments and obtain a bunch of prefixes. We then generate 20 completions for each prefix and examine the completion. We give a positive score to a prefix if there is at least one correct completion. Please refer to Appendix~\ref{app:dataset} for more details.


\begin{table}[ht]
\centering
\caption{Experiment results for DQO augemented by process scores. With process rewards, when using greedy decoding, the performance of DQO is further enhanced by 0.53\% on GSM8K and 0.32\% on MATH. The performance when doing sampling also increase on GSM8K and maintains almost the same on MATH.} 
\vspace*{0.1in}
\begin{tabular}{l | c | c | c | c }
\toprule
Dataset             & \multicolumn{2}{c|}{GSM8K} & \multicolumn{2}{c}{MATH}  \\
Decoding       & Greedy  & Sample  & Greedy   & Sample \\
\midrule
Gemma & 39.65 & 37.89$_{\pm 1.02}$   & 17.04  & 16.14$_{\pm 0.21}$               \\
\midrule
DQO   & 64.51 & 64.00$_{\pm 0.37}$  & 24.90   & \textbf{24.84$_{\pm 0.29}$}   \\
DQO$_{\text{w/ PS}}$  & \textbf{65.04}  & \textbf{64.55$_{\pm 0.66}$}  & \textbf{25.22}      & 24.70$_{\pm 0.38}$  \\
\bottomrule
\end{tabular}
\label{tab:result-gemma-prm}
\end{table}

We conduct the experiments on Gemma, and the results are summarized in Table~\ref{tab:result-gemma-prm}. Equipped with our estimated process scores, we see a further improvement. Specifically, on GSM8K, using our process scores further increases the performance by 0.53\% for greedy decoding and 0.55\% for sampling. On MATH, process scores also boost the model's performance by a further 0.32\% when doing greedy decoding and maintains almost the same to DQO without process reward for sampling. The results imply that DQO can be further enhanced by utilizing process scores. 

\section{Conclusion}

In this work, we propose DQO, an offline reinforcement learning algorithm for enhancing the language model's ability in multi-step reasoning. Compared to previous online methods like PPO, the offline nature of DQO bypasses the requirement of an extra reward model and online sampling during training. Previous offline methods usually formulate the LLMs' responding process as a bandit problem, which usually fails to capture the implicit long-horizon and multi-step nature of those tasks requiring a long chain of thought. In contrast, DQO frames the tasks as a Markov decision process and further employs a soft actor-critic framework to learn the $V$-function and the $Q$-function, which is directly parameterized by the language model. To verify the effectiveness of DQO, we conduct extensive experiments on math-problem-solving datasets, and empirical results show that DQO outperforms all our baselines.

\appendix

\section{Additional Experiment Details}

\subsection{Dataset Construction and Evaluation}\label{app:dataset}

\begin{figure*}[ht]
    \centering
    \includegraphics[width=.98\textwidth]{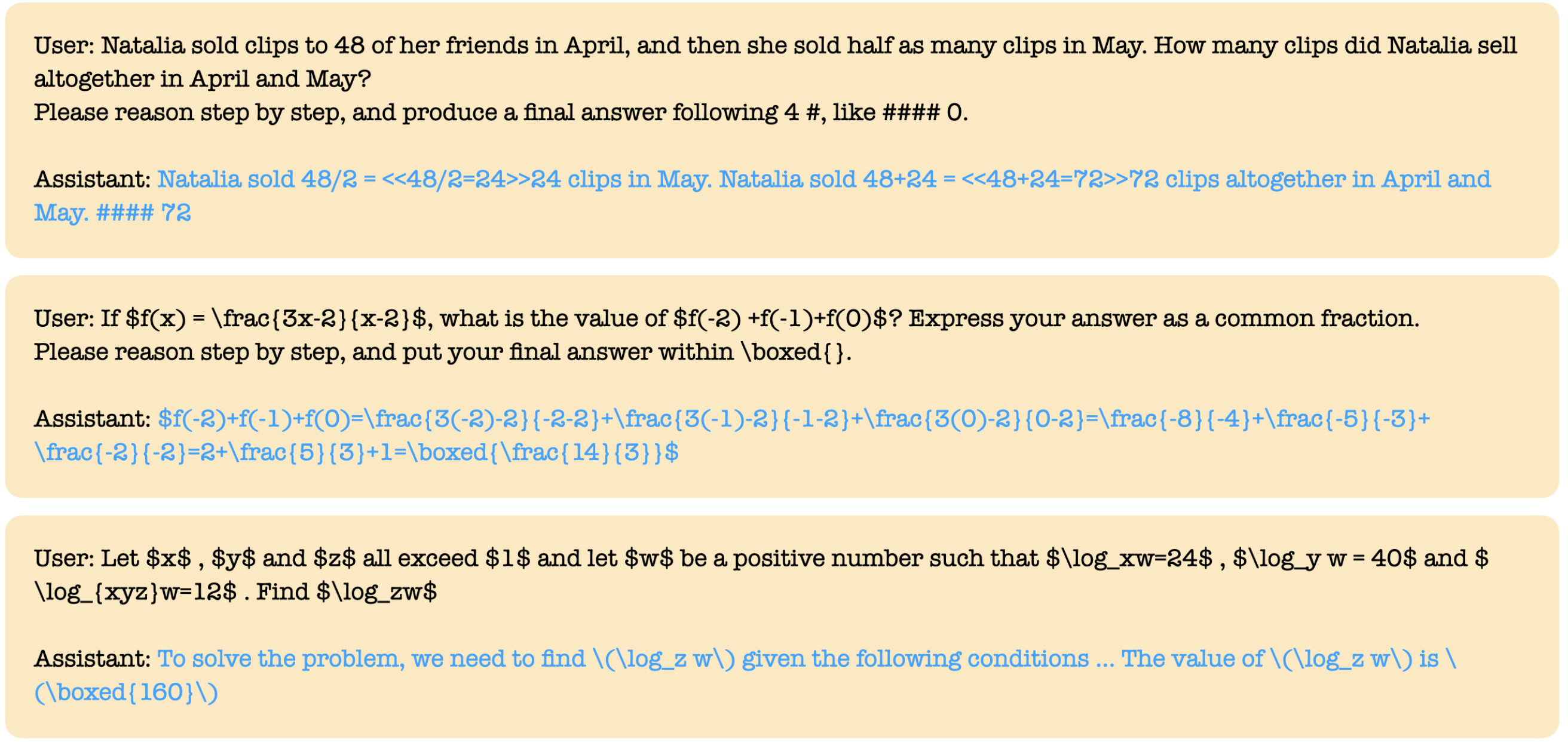}  
    \caption{Examples of prompting LLMs with math problems from different dataset. The problems, wrapping in the employed prompt template, from top to bottom, are from GSM8K, MATH and Numina respectively. Here texts in black is the prompt we feed to LLM and texts in light blue are sample answers. }
    \label{fig:prompt-template}
\end{figure*}
\begin{figure*}[ht!]
\centering   

\subfigure[Gemma-MATH]{\label{fig:gemma-math-stat}\includegraphics[width=0.32\textwidth]{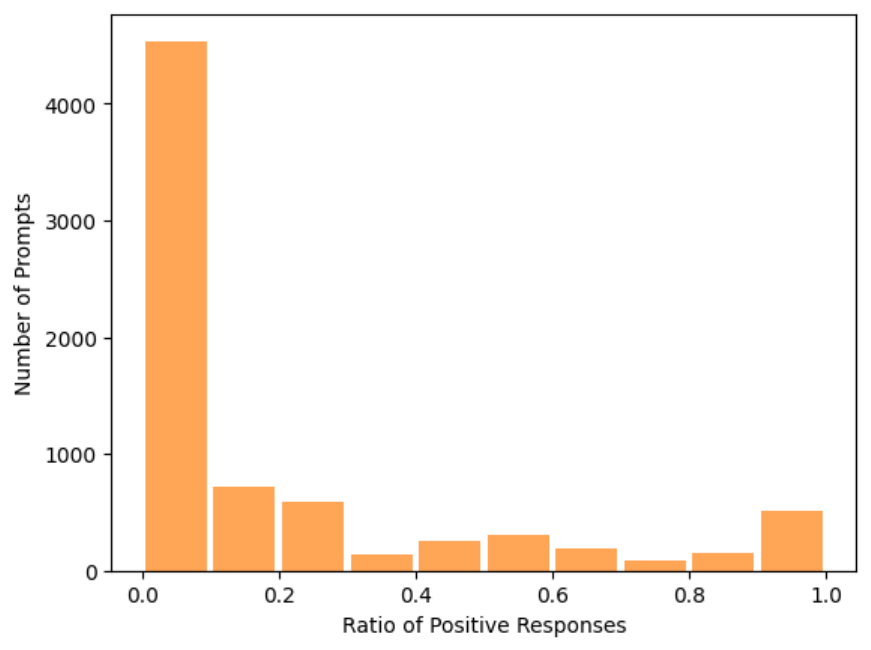}}
\subfigure[Gemma-GSM8k]{\label{fig:gemma-gsm8k-stat}\includegraphics[width=0.32\textwidth]{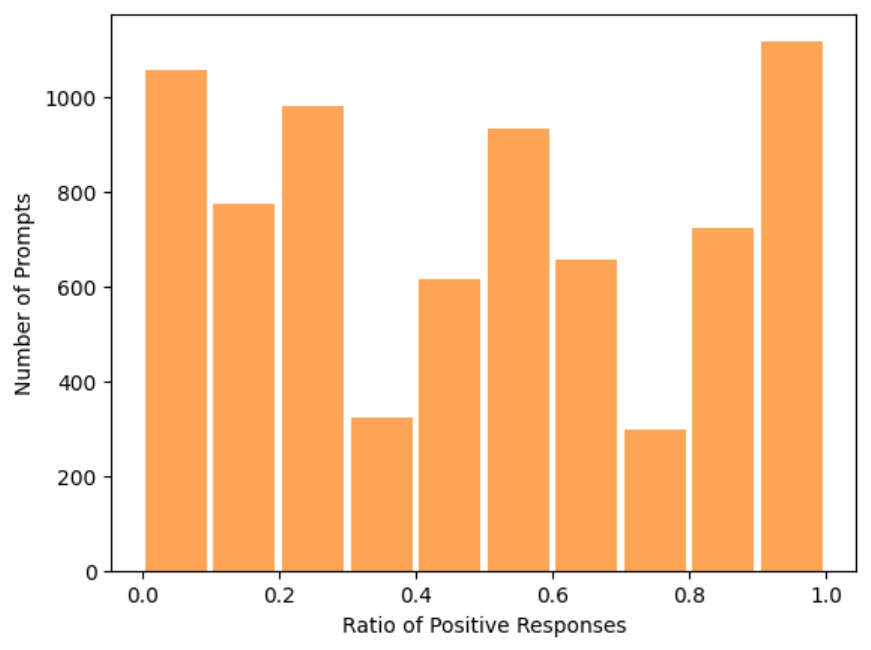}}
\subfigure[Qwen2.5-QwQ-Numina]{\label{fig:qwq-base-aops-stat}\includegraphics[width=0.32\textwidth]{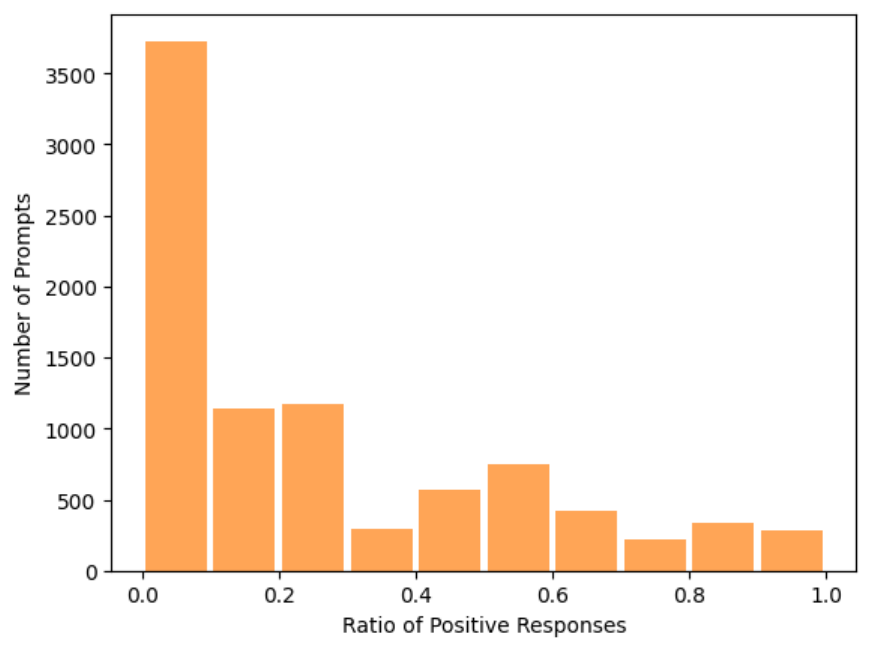}}
\vspace*{-2mm}
\caption{The distribution of positive and negative prompts in each training set. The x-axis is the proposition of positive responses over all responses given a specific problem and y-axis is the count of prompts. It can be seen that the generations of most prompts are highly unbalanced especially for Gemma-MATH and Qwen2.5-QwQ-Numina (negative responses dominates most of the problems). While hard to utilized by reject sampling and DPO, DQO can still make use of these responses.}
\label{fig:trainset-stats-acc}
\end{figure*}
\begin{figure*}[ht!]
\centering   
\subfigure[Gemma-MATH]{\label{fig:gemma-math-length}\includegraphics[width=0.32\textwidth]{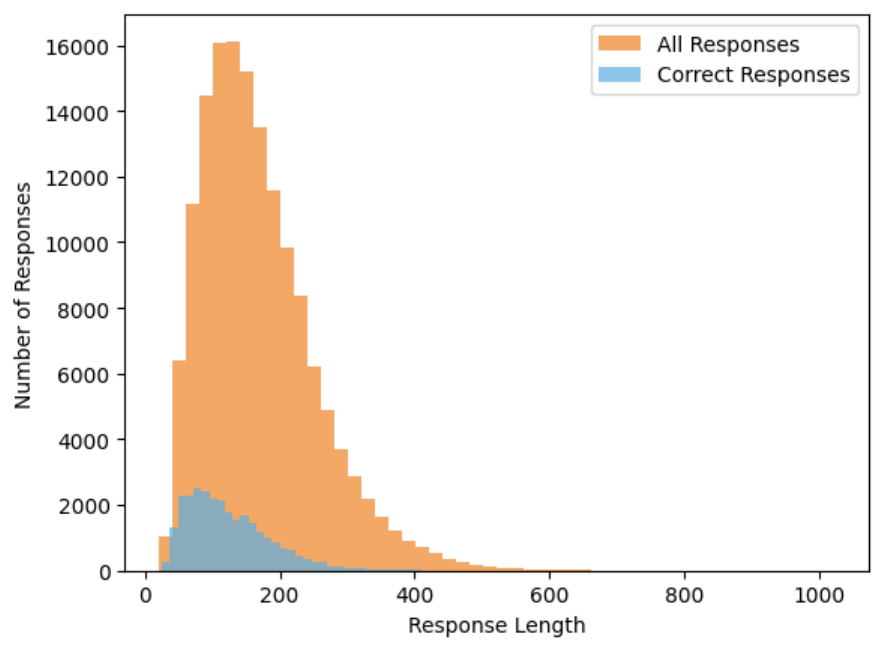}}
\subfigure[Gemma-GSM8k]{\label{fig:gemma-gsm8k-length}\includegraphics[width=0.32\textwidth]{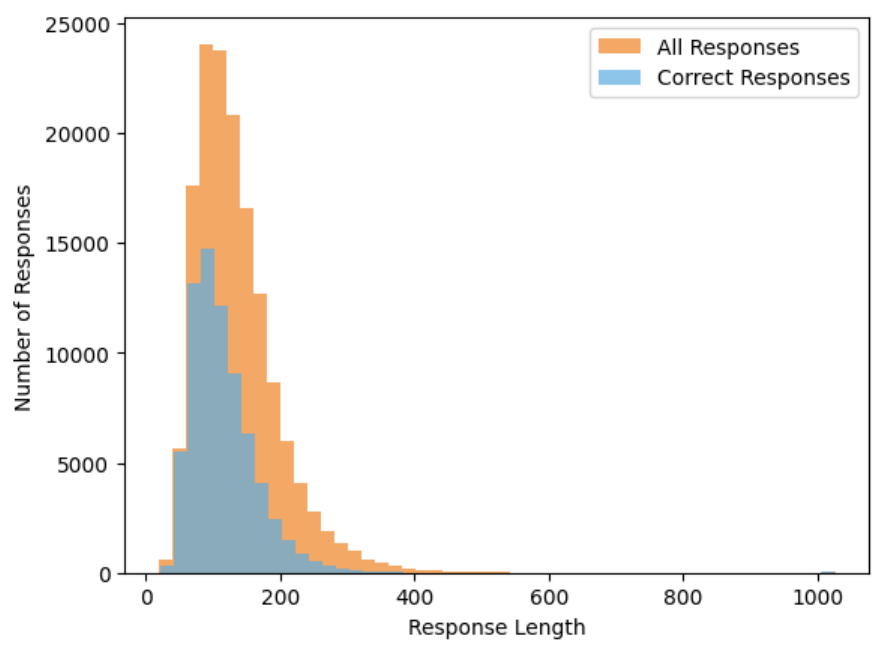}}
\subfigure[Qwen2.5-QwQ-Numina]{\label{fig:qwq-base-aops-length}\includegraphics[width=0.32\textwidth]{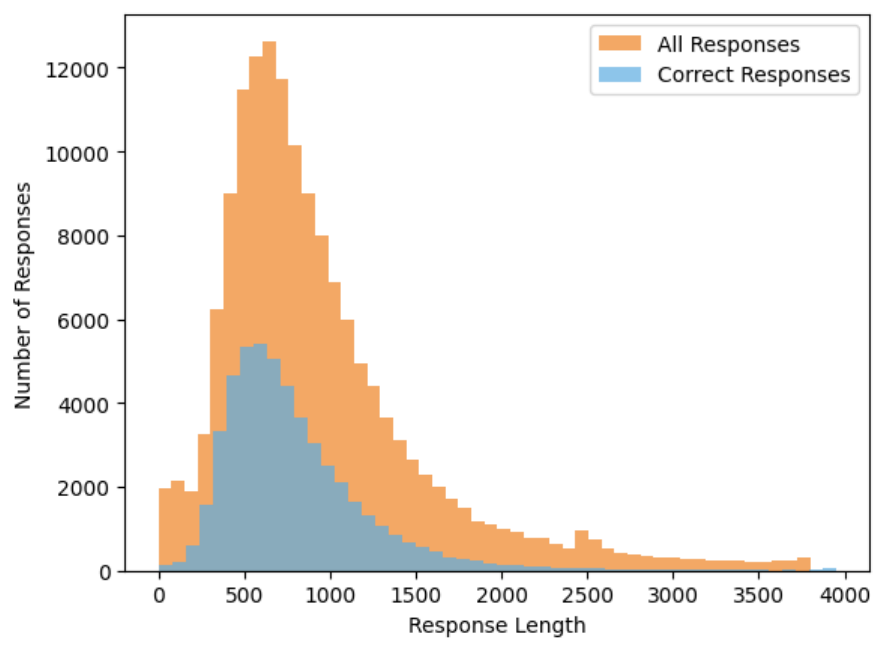}}
\vspace*{-2mm}
\caption{The length distribution of all generated responses and positive responses. The x-axis is the length of responses (number of tokens) and y-axis is the number of responses. For Qwen2.5-QwQ, we only count all responses that have not been trimmed to the maximum sequence length of 4096 tokens. We observe that the responses generated by Qwen2.5-QwQ is significantly longer than those generated by Gemma.}
\label{fig:trainset-stats-length}
\end{figure*}

\paragraph{Dataset Construction } In this section, we provide details about our training set construction. For each dataset, we sample 20 responses for every problem in it. For generation, we employ format guide on MATH and GSM8K to make the language models' generation follow some specific format and directly feed the problems in Numina and AIME24 to LLM. Please refer to Figure~\ref{fig:prompt-template} for the prompt templates we used.
We then wrap the prompts with the default chat template of the tokenizers. We set the sampling parameters to \texttt{top\_p}=0.9, \texttt{top\_k}=16, \texttt{threshold}=0.01 and \texttt{temerature}=0.9. We then follow the evaluation process and score each generated response. 
The distribution of then amount of positive and negative responses of each problem are shown in Figure~\ref{fig:trainset-stats-acc}. We also count the length distribution of all responses and all correct responses, which are shown in Figure~\ref{fig:trainset-stats-length}. We then construct our trainset with these annotated responses.
For reject sampling, we collect all correct responses as the training target. For DPO, we first pair up all positive and negative responses and then randomly sample all possible pairs to make the size of the DPO training dataset approximately half of DQO, which means that the dataset contains a similar number of trajectories. The size of training datasets are summarized in Table~\ref{tab:dataset-size}.

\begin{table}[ht]
\centering
\caption{The size of datasets for all of our baselines and DQO. We ensure that the size of DPO training set is at least half of the training set of the training set for DQO. This guarantees that the number of trajectories in DPO dataset is no less than the number of trajectories in the dataset of DQO for a fair comparison.}
\vspace*{0.1in}
\begin{tabular}{l | c | c | c | c | c | c}
\toprule
  Model      &    Dataset & Size of Trainset   & SFT  & RS & DPO   & KTO/DRO/DQO    \\
        \midrule
        & MATH & 7500  & 7500 & 94117   & 96598 & 150000 \\
\multirow{-2}{*}{Qwen2}  & GSM8K & 7473 & 7473 & 94889  & 87996 & 149460 \\
\midrule
        & MATH  & 7500 & 7500 & 28295  & 93314 & 150000 \\
\multirow{-2}{*}{Gemma} & GSM8K & 7473 & 7473 & 46062  & 79523 & 149460 \\
\midrule
Qwen2.5 & Numina & 8919 & - & 44506 & 121332 & 178380  \\
\bottomrule
\end{tabular}
\label{tab:dataset-size}
\end{table}

\begin{figure*}[ht!]
    \centering
    \includegraphics[width=.9\textwidth]{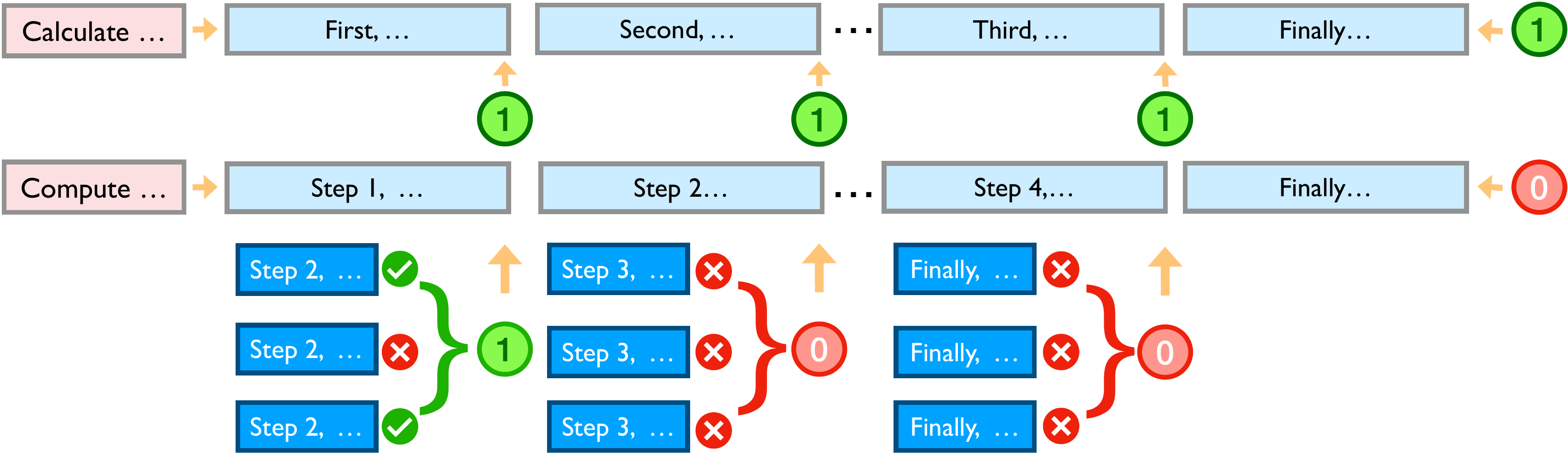}  
    \caption{A visual demonstration of our process reward construction. We split all the responses to segments. For correct responses we assign all process reward to one. For negative responses, we start from each prefix and generate 20 samples. We then find the longest prefix where the best of 20 samples is correct and assign all the process rewards before to 1. }
    \label{fig:prm-construct}
\end{figure*}

\paragraph{Process Score Construction } We consider using an empirical passing rate to estimate the quality of a given response prefix. Specifically, given a prompt string $x$, for each failed response $y$, we first split the response into several segments $y[0:n]$, where $n$ is the number of segments and we use $y[0:i]$ to denote the concatenation of first $i$ segments. Beginning from $i=n-1$, we randomly sample 20 trajectories given prefix $\mathrm{contat}(x,y[0:i])$. If there is at least one correct completion, we assume that the reasoning process in $y[0:i]$ is correct and all the process rewards for the previous step will be set to $1/n$. We combine these process reward scores with the original rewards. We also refer the readers to Figure~\ref{fig:prm-construct} for a visual illustration.

\paragraph{Evaluation } For each problem in test set, we employ the same formatting instructions as constructing the training set. We considered two decoding strategy, greedy decoding and sampling. When sampling, we set the sampling parameters to \texttt{top\_p}=0.9, \texttt{top\_k}=16, \texttt{threshold}=0.01 and \texttt{temerature}=0.7, and sample for 5 times to settle the randomness. We adopt the implementation of Qwen2.5-Math\footnote{https://github.com/QwenLM/Qwen2.5-Math/tree/main/evaluation} to score the responses. Specifically, in GSM8K dataset, we use regular expression to extract the prediction (which should appear after the pattern "\#\#\#\#") from the generated response. For MATH and AIME dataset, we first find the formatting "\textbackslash boxed" and then parse the \LaTeX \ expression after the formatting to obtain the prediction. We follow a similar procedure to obtain the ground-truth answers from the reference solutions. A prediction is considered correct only when the prediction and ground-truth answer are mathematically equivalent.

\subsection{Additional Training Details}\label{app:baseline-hyper}

We conducted all our experiments on $8 \times$ NVIDIA A100 GPUs with approximately 80G memories. For Qwen2 and Gemma, on both MATH dataset and GSM8K dataset, it take approximately 1 hour for training with reject sampling, 4 hours for training with DPO and KTO, 6 hours for training with DRO and 8 hours for training with DQO. For Qwen2.5-QwQ, it takes approximately 1 hour for training with reject sampling, 8 hours for training with DPO and KTO, 10 hours for training with DRO and 12 hours for training with DQO.


\paragraph{SFT and Reject Sampling } For SFT and reject sampling, we select the best learning rate from \{2e-5, 1e-5, 5e-6, 1e-6\} and the best epoch from \{1,2,3\}. For SFT, the final learning rate is set to 2e-5 for Qwen and 5e-6 for Gemma. For reject sampling, the final learning rate is set to 2e-5 for Qwen2, 1e-6 for Gemma and 2e-5 for Qwen2.5-QwQ. We set global batch size of 8 and therefore global batch size to 64. We trained the model for 3 epoches for both SFT and reject sampling.

\paragraph{DPO and KTO } For DPO, we tried $\beta$ from \{0.1, 0.01\} and learning rate from \{5e-7, 1e-7, 5e-8\} and select the hyperparameter set that yields the best performance. Specifically, for Qwen2, Qwen2.5 and Gemma we set $\beta$ to $0.1$. The learning rate is set to 5e-8 on Gemma and 1e-7 on both Qwen2 and Qwen2.5-QwQ. We set the local batch size to 8 and therefore global batch size to 64. We train the model for $1$ epoch. As recommended by the original paper of KTO, We adopt the same set of hyperparameters of DPO to train KTO. 

\paragraph{DRO and DQO } For both DRO and DQO, we try the KL regularization parameter $\beta$ from $\{0.01, 0.03, 0.1, 0.3, 1\}$, learning rate for policy updating from \{5e-7, 1e-7, 5e-8\} and value updating from \{5e-6, 1e-6\}. We then select the set of parameters that yields the best results. The final parameter for both DRO and DQO is $\beta=0.03$ and we set learning rate to 5e-7 for Qwen2, Qwen2.5-QwQ and 1e-7 for Gemma. We set the local batch size to 32 and therefore the global batch size to 256. We train the model for a maximum of 5 epoches and select the best checkpoints on the training curve for evaluation. The final DRO checkpoints is the checkpoint after the first training epoch. For Qwen2 and Gemma, we pick the checkpoint at the end of second epoch as the final checkpoint and pick the checkpoint after the first epoch for Qwen2.5-QwQ.

\section{Additional Experiment Results}

\subsection{Results on Qwen2-7B-Instruct}

We also conduct experiments on \textit{Qwen2-7B-Instruct\footnote{https://huggingface.co/Qwen/Qwen2-7B-Instruct}}~\citep{yang2024qwen2} (Qwen2) and summarize our results in Table~\ref{tab:result-qwen2}. The results show that all methods improve the performance of the base models by a significant margin. Particularly, on GSM8K, DQO improves the performance from 72.77\% to 87.95\% for greedy generation and 60.77\% to 85.13\% for sampling. This improvement is comparable with DPO and surpasses DRO and other baselines by a margin of 0.70\% for sampling and 1.22\% for greedy generation.
On MATH, we also see a significant performance improvement from DQO. As for greedy decoding, the performance of DQO, while comparable with DPO and DRO, surpasses all other baselines by a margin of 1.64\%. As for sampling, DQO reaches a performance of 49.36\%, which surpasses the performance of the best baseline method DPO by a margin of 1.12\%. Moreover, DQO also improve the performance on AIME24 from 8.55\% to 10.56\% and surpasses all the baselines by a margin of more than 0.59\%. These results indicate that DQO achieves a comparable performance of DPO and surpasses other baselines by a considerable margin. These results are consistent with the results from Qwen2.5-QwQ and Gemma, demonstrating a superior performance of DQO over other baselines.

\begin{table*}[hbt!]
\centering
\caption{Experiment results for \textit{Qwen2-7B-Instruct} model. We use \textbf{bold} for the best and \underline{underline} for the second best. DQO significantly improves the base model's performance. This improvement surpass all the baselines when doing greedy decoding. As for sampling, DQO is comparable to DPO and surpass all other baselines. } 
\vspace*{0.1in}
\begin{tabular}{ l | c | c | c | c | c}
\toprule
Dataset                 & \multicolumn{2}{c|}{GSM8K}  & \multicolumn{2}{c|}{MATH} & \multirow{2}{*}{AIME24}  \\
Decoding       & Greedy   & Sample    & Greedy     & Sample & \\
\midrule
Qwen2-7B-Instruct & 76.19 & 60.77$_{\pm 1.62}$ & 53.74 & 50.27$_{\pm 0.45}$ & 8.55 \\
\midrule
Qwen2-7B-Instruct + SFT & 85.06 & 84.06$_{\pm 0.66}$ & 54.98 & 54.52$_{\pm 0.44}$ & 9.66 \\
Qwen2-7B-Instruct + RS & 84.15 & 84.43$_{\pm 0.59}$ & 57.50 & 54.76$_{\pm 0.49}$ & 8.11 \\
Qwen2-7B-Instruct + DPO & 85.35 & \textbf{85.67$_{\pm 1.01}$} & 58.68 & 54.98$_{\pm 0.41}$ & \underline{9.97} \\
Qwen2-7B-Instruct + KTO & 86.35 & 83.52$_{\pm 0.64}$ & \underline{58.84} & \underline{55.05$_{\pm 0.19}$} & 9.22 \\
Qwen2-7B-Instruct + DRO & \underline{86.73} & 82.56$_{\pm 0.48}$ & 58.50 & 54.08$_{\pm 0.25}$  & 8.44 \\
\midrule
\rowcolor{LightCyan} Qwen2-7B-Instruct + DQO & \textbf{87.95} & \underline{85.13$_{\pm 0.47}$} & \textbf{59.14}  & \textbf{56.23$_{\pm 0.32}$}  & \textbf{10.56}  \\
\bottomrule
\end{tabular}
\label{tab:result-qwen2}
\end{table*}

\subsection{Additional Case Studies}\label{app:case-study}

In this section, we select some additional cases to provide a more comprehensive implication of how the value model introduced in DQO provides process supervisions. Due to space limit, we select prompts from GSM8K and MATH datasets, which include relatively simple examples, and all the responses are generated by Gemma base model. Please refer to Table~\ref{table:add-case-math} and Table~\ref{table:add-case-gsm8k} for details. We see that in both cases, the learned $V$-values reflect the correctness of each reasoning step in the response. 


\begin{table*}[t!]
\centering
\Huge
\caption{Another prompt-response pair from the MATH training set. The response makes a mistake at its third statement. Correspondingly the predicted value dropped to significantly after the mistake.}
\vspace*{0.1in}
\resizebox{0.95\columnwidth}{!}{
\begin{tabular}{l | l | c | c}
\toprule
&  \textbf{Text} & \textbf{Process Value} & \textbf{Correctness} \\ \midrule
\textbf{Prompt} &  
\parbox[t]{25cm}{
Find the minimum value of
\textbackslash frac\{x\string^2 + 7\}\{\textbackslash sqrt\{x\string^2 + 3\}\} over all real numbers \$x\$ . \\
Please reason step by step, and put your final answer within \textbackslash boxed\{\}. \vspace{15pt}
}
 & - & -\\ \midrule
\textbf{Response} 
& \colorbox{lightgreen}{\parbox[b]{25cm}{\texttt{$$\frac{(x^2 + 7)}{\sqrt{x^2 + 3}}.$$} \\}} & 0.2474 & \textcolor{Green}{\ding{51}} \\ 
  & \colorbox{lightgreen}{\parbox[b]{25cm}{ \texttt{$$ \text{ Let } y = \sqrt{x^2 + 3} \text{, so that } x^2 + 3 = y^2$$} \\}}
     & 0.1593 &  \textcolor{Green}{\ding{51}} \\
  & \colorbox{lightred}{\parbox[b]{25cm}{ \texttt{$$ \frac{x^2 + 7}{y} = \frac{y^2 - 3 + 7}{y} = y + \frac{7}{y}$$} \\}}
     & 0.0101 & \textcolor{Red}{\ding{55}}  \\
  & \colorbox{lightred}{\parbox[b]{25cm}{ \texttt{$$\text{ Apply AM-GM inequality for } y + \frac{7}{y} :$$} \\}}
     & -0.0094 & \textcolor{Red}{\ding{55}}  \\
  &  \colorbox{lightred}{ \parbox[b]{25cm}{ \texttt{$$ y + \frac{7}{y} \geq 2\sqrt{y \left(\frac{7}{y}\right)} = 2\sqrt{7}$$}\\} }
     & 0.0689 & \textcolor{red}{\ding{55}}  \\
  &  \colorbox{lightred}{\parbox[b]{25cm}{ \texttt{$$\text{ So the minimum value is } 2\sqrt{7}.$$}\\} }
     & 0.0981 & \textcolor{red}{\ding{55}}  \\
  & \colorbox{lightred}{ \parbox[b]{25cm}{ \texttt{$$\boxed{ \text{ Minimum value = } 2\sqrt{7} }$$}\\} } & - & \textcolor{red}{\ding{55}}  \\ 
\bottomrule
\end{tabular}
}
\label{table:add-case-math}
\end{table*}

\begin{table*}[htb!]
\centering
\Huge
\caption{Another prompt-response pair from the GSM8K training set. The response gives the correct answer in this case and the predicted $V$-values gradually increase, which reflect the correctness of the reasoning.}
\vspace*{0.1in}
\resizebox{0.95\columnwidth}{!}{
\begin{tabular}{l | l | c | c}
\toprule
&  \textbf{Text} & \textbf{Process Value} & \textbf{Correctness} \\ \midrule
\textbf{Prompt} &  
\parbox[t]{25cm}{
Tim gets 6 hours of sleep 2 days in a row.  To make up for it he sleeps 10 hours the next 2 days.  How much sleep did he get? \\
Please reason step by step, and produce a final answer following 4 '\#', like '\#\#\#\# 0'.. \vspace{15pt}
}
 & - & -\\ \midrule
\textbf{Response} & \colorbox{lightgreen}{\parbox[b]{25cm}{\texttt{**Step 1: Calculate the sleep Tim got in the first 2 days.**} \\}} & 0.5859 & \textcolor{Green}{\ding{51}} \\ 
  & \colorbox{lightgreen}{\parbox[b]{25cm}{ \texttt{6 hours/day * 2 days = 12 hours} \\}}
     & 0.8672 &  \textcolor{Green}{\ding{51}} \\
  & \colorbox{lightgreen}{\parbox[b]{25cm}{ \texttt{**Step 2: Calculate the sleep Tim got in the next 2 days.**} \\}}
     & 0.6797 & \textcolor{Green}{\ding{51}}  \\
  & \colorbox{lightgreen}{\parbox[b]{25cm}{ \texttt{10 hours/day * 2 days = 20 hours} \\}}
     & 0.8008 & \textcolor{Green}{\ding{51}}  \\
  &  \colorbox{lightgreen}{ \parbox[b]{25cm}{ \texttt{**Step 3: Add the sleep from both sets of days.**}\\} }
     & 0.7695 & \textcolor{Green}{\ding{51}}  \\
  &  \colorbox{lightgreen}{\parbox[b]{25cm}{ \texttt{12 hours + 20 hours = 32 hours}\\} }
     & 1.0625 & \textcolor{Green}{\ding{51}}  \\
  & \colorbox{lightgreen}{ \parbox[b]{25cm}{ \texttt{\#\#\#\# 32}\\}} & - & \textcolor{Green}{\ding{51}}  \\ 
\bottomrule
\end{tabular}
}
\label{table:add-case-gsm8k}
\end{table*}

\bibliographystyle{ims}
\bibliography{reference}

\end{document}